\documentclass{article} 
\usepackage{nips15submit_e,times}
\usepackage{fullpage}
\usepackage{hyperref}
\usepackage{url}
\usepackage{graphicx}
\usepackage{epsf, verbatim}
\usepackage{amssymb,amsmath}    
\usepackage{gensymb}
\usepackage{array}
\newcolumntype{M}[1]{>{\centering\arraybackslash}m{#1}}
\newcolumntype{N}{@{}m{0pt}@{}}

\title{Classification-based RNN machine translation using GRUs}

\author{
Ri Wang \\
\texttt{rtwang@uwaterloo.ca} \\
\And
Maysum Panju \\
\texttt{mhpanju@uwaterloo.ca} \\
\And
Mahmood Gohari \\
\texttt{mgohari@uwaterloo.ca} \\
}

\nipsfinalcopy 

\begin{document}

\bibliographystyle{plain}
\maketitle

\begin{abstract}
We report the results of our classification-based machine translation model, built upon the framework of a recurrent neural network using gated recurrent units. Unlike other RNN models that attempt to maximize the overall conditional log probability of sentences against sentences, our model focuses a classification approach of estimating the conditional probability of the next word given the input sequence. This simpler approach using GRUs was hoped to be comparable with more complicated RNN models, but achievements in this implementation were modest and there remains a lot of room for improving this classification approach.
\end{abstract}

\section{Introduction}
Machine translation is an increasingly important field of study, as the need for natural language translation is rapidly rising in a world that is growing increasingly multilingual and interconnected. At the same time, machine translation is an incredibly difficult task whose problems span a range of linguistic fields and is difficult even for human translators. Previous efforts have been centered on statistical machine translation, but the continuous and exponential increase in computing power and advances in neural network training have enabled researchers to use recurrent neural networks to approach this problem.

Currently, neural networks such as LSTMs and bidirectional RNNs are trained for this purpose. However, methods described in many of these research papers are relatively complex and frequently non-intuitive. The approach taken by feed-forward neural network researchers, however, involves using the neural network as a classifier to predict probabilities, and is a simpler and more attractive framework to build upon. In this work, we attempt to remove the common constraint in feed-forward networks that require input sequences to have a fixed-length. Instead, we propose to build a relatively simple GRU neural network model, which both allows variable-length input and is able to act as a classifier applied to the task of machine translation.

\section{Background and Related Work}
\label{BackgroundAndRelatedWork}

The approach for translation using recurrent neural networks was mainly based on the model given by Sutskever et. al. \cite{sutskever2014sequence} of recursively inputting output into input to generate a translation. In their paper, they trained pairs of long short-term memory models (LSTM's): one LSTM served as a feature extractor to convert a variable-length sentence into a fixed-length vector such that sentences with similar meanings are mapped close together, and the other LSTM takes the word vectors as hidden states and proceeds to output words one at a time, incrementally building up the translation. It was not entirely clear how this pair of LSTM training was done in the context of back-propagation, and we decided to focus on their alternative model of having a single RNN to perform both feature extraction and translation. The model we used was also very closely related to that of Cho et al. \cite{cho2014learning}, which introduced the GRU and applied it as an RNN encoder-decoder machine translation model as an alternative to LSTM. 

However, the classification approach taken here is closer related to previous work done by Schwenk  \cite{schwenk2012continuous}, with feed-forward neural networks that take in a fixed input sentence and output the probability of a sequence of words occurring next, as well as the work of Devlin et al., who only predicted the next word \cite{devlin2014fast}. The difference in our approach is that the GRU used here can take in a variable-length sentence, and is much more flexible than the models by Schwenk and Devlin et al., which were restricted to only accepting input of a pre-specified fixed length.

\section{Basic Framework of the Learning Models}
\label{Setup}

The key ingredient for training a machine translation model is a large corpus of sentences in the source language that has been correctly translated into sentences in the target language. For a powerful translator, this training corpus must be very large and the translations must be perfectly accurate.

There are other specifications and requirements that need to be met as well. The model requires both source and target languages to draw from fixed vocabulary sets; in this work, we restrict both the input and output vocabulary sets to each consist of 80,002 distinct tokens, representing the 80,000 most frequently appearing words in the training set for each language, along with 2 additional tokens added to represent the not-found token and the end-of-sentence token. To decrease the number of unique words, all words were converted into lowercase, but were neither stemmed nor lemmatized. This is because grammatical inflections in words are very important to overall translation quality, as they carry heavy semantic value that needs to be preserved during translation.

Prior to use in the model, words were converted into word vectors using a standard continuous word-to-vector embedding \cite{mikolov2013distributed}. The translation model would eventually be trained to recognize that semantically similar words should be weighted to carry similar representations, and using word embeddings that already indicate semantic similarity from the outset should reduce a lot of training effort on the part of the translation model. Furthermore, we expect higher quality word embedding from specialized representation models, such as GloVe \cite{pennington2014glove}, than any word embedding built into the translation neural network such as the one used by Cho et al. \cite{cho2014learning},  because they are trained on a much larger dataset and their linguistic capabilities have been proven to be very strong.

The training was based on stochastic gradient descent with mini-batches of 128 sentence pairs each. To take full advantage of GPU parallelization, batches were prepared so that within each batch $B_i$, all of the source-language input sequences consisted of the same number of tokens $F_i$, and all of the target-language output sequences consisted of the same number of tokens $E_i$, although it was not necessary that $F_i$ be the same as $E_i$ for any given batch. This condition was enforced to prevent part of the batch dropping out of the back-propagation gradient calculations midway, in the event that the some sequences reached completion before other sequences in the batch. If there were not enough sentence pairs in the training set corresponding to a given pair of lengths $(F_i, E_i)$ to fill a minibatch of size 128, then the entire batch $B_i$ was dropped from the dataset altogether; this measure filtered out a large number of small batches, which had been observed to learn much less than full-size batches, despite taking almost the same amount of time to train. Additionally, both $F_i$ and $E_i$ were hard-capped with a maximum of 50, because the neural network would have trouble handling translations between lengthy sentences of longer than 50 tokens in either the source or target language.

\section{Model}

In this section, we describe the procedure that the deep learning model is used to perform machine translation, once the preprocessing and setup from Section~\ref{Setup} has been implemented.

\subsection{Automatic Machine Translation}
\label{AutoTranslate}

After the input text in the source language has been represented as sequences of word embedding vectors of fixed length, our translation model works as follows:

For a given sequence of input word vectors ($x_1,\dots,x_F$) encoding a sentence in the source language, as well as the sequence of output word vectors ($y_1, \ldots, y_k$) encoding the tokens in the translated sentence in the target language so far, the neural network estimates the conditional probability of the word vector representation of the next word in the translated sequence $y_{k+1}$. Thus, part of the input sequence to the neural network would be a complete sequence of source-language word vectors and the remainder of the input would be a sequence in progress consisting of target-language word vectors. 

The output vector of this neural network is a vector of length 80,002, with components corresponding to each of the words in the target language vocabulary, along with the two special tokens representing not-found and end-of-sentence. The entry at position $i$ in this vector, for $1 \leq i \leq 80,002$, is a real-valued number between 0 and 1 representing the probability that the $i$th word in the target language vocabulary set, $v_i$, is the next word in the translation sequence.

If the neural network is treated as a function $f$, then its operation is defined as:

\[f((x_1, \ldots, x_F), (y_1, \ldots, y_k) ) = [P(y_{k+1} = v_i|(x_1, \ldots, x_F), (y_1, \ldots, y_k))]_{i = 1, \ldots, 80,002}  \]

As the translation is built up word-by-word, the neural network can be treated as a classifier that takes in a sequence of input vectors and predicts the ``class'', or vocabulary word, that should appear next in the sequence. 

Finally, to calculate the conditional probability of obtaining the output translation sentence $y_{1},\dots,y_{E}$ given the input sentence word sequence $x_{1},\dots,x_{F}$ for the purpose of translation, the probability may be decomposed into a product of conditional probabilities that the neural network is able to compute:

\[P((y_{1},\dots,y_{E})|(x_{1},\dots,x_{F}) = 
\prod_{k=0}^{E-1} P(y_{k+1}|(x_1, \ldots, x_F), (y_1, \ldots, y_k))\]

The predicted translation is then the sentence with the highest overall conditional probability, where the probabilities may be obtained by calculating each term in the above product using the neural network. At each stage, the input is the source sentence sequence $(x_{0},\dots,x_{F})$ and the incrementally built-up target sequence $(y_{0},\dots,y_{k})$.

\subsection{The neural network model: GRUs}

The main contribution of deep learning to the task of automatic machine translation, in the approach taken by this work, is the computation of the conditional probabilities for the next translated word given the progress of the translation so far. This approach has been applied using various specializations of neural networks for the task, including long short-term memory networks \cite{hochreiter1997long} and bidirectional recurrent neural networks \cite{schuster1997bidirectional}. In this work, we explore the possibility of using gated recurrent units in the model.

A gated recurrent unit (GRU) was initially proposed by Cho et. al. \cite{cho2014learning} for their translation task. The idea behind this approach is to keep the current content at each step in the neural network, and add the new contents to that, while forgetting any past information that no longer adds additional information to the present state. Unlike the LSTM approach that define a separate memory cell, the GRU resets a gate and stores only the relevant new content to the unit. The two approaches of GRU and LSTM in handling variable length inputs are presented for contrast in Figure 1.

\begin{figure}[h]
  \centering
  \includegraphics[width=0.9\textwidth]{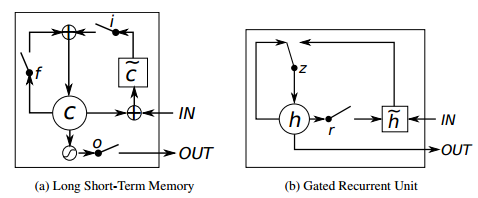}
  \caption{Long-Short Term Memory (left) and Gated Recurrent Units (right) gating. $c$ and \(\tilde{c}\) stand for the memory cell and the new memory cell content in LSTM. $r$ and $z$ are the reset and update gates, and $h$ and \(\tilde{h}\) are the activation and the candidate activation in GRU. Figure copied from Chung, J, et al.: Empirical evaluation of gated recurrent neural networks on sequence modeling \cite{chung2014empirical}.}
\end{figure}

Given an input sequence \(x=(x_1,x_2,...,x_T)\), the standard RNN computes the activation by
\[h_t=a(Wx_t+Uh_{t-1})\]
where \(a\) is a smooth activation function such as hyperbolic tangent function. The GRU computes an update gate as 
\[z_t^j=\sigma^j(W_zx_t+U_zh_{t-1})\]
This update gate is the sum of current state at \(t\) and the recent state which controls the amount of previous memory to be kept. In the LSTM, the update gate has an additional component that is the memory cell at time \(t\).  Similar to the update gate, a reset gate is computed based on the current input vector and hidden state, but with different weights:
\[r_t^j=\sigma^j(W_vx_t+U_vh_{t-1})\]
Now the activation is computed as 
\[\tilde{h_t}^j=\tanh^j\left(Wx_t+U(r_t\odot h_{t-1})\right)\]
where \(\odot\) represents pointwise multiplication. If the reset gate is zero, then the previous state will be ignored and only the current input is used for updating. The final output at time \(t\) is computed as a weighted average of \(\tilde{h}_t\) and \(h_t\)
\[h_t=z_t\odot h_{t-1}+(1-z_t)\odot\tilde{h}_t\]

\section{Training Methods and Implementation}
\label{TrainMethods}

For this work, we made use of the publicly available Europarl\footnote{The European Parliament dataset, restricted largely to political topics and certain styles of speaking, is not representative of speech as a whole, which means the model might not be able to generalize well under other circumstances. However, it is not easy to obtain high quality translation data, and we have to make do with what is available.} corpus \cite{koehn2005europarl}, with French as the source language and English as the target language. This dataset contains about two million instances of English and French sentence pairs across a large variety of sentence lengths. Translation was done from French to English because this report is intended for an English-speaking audience, and a model with English-language output allows us to assess the quality of the translation directly and independently. Furthermore, this approach distinguishes our work from that of many other papers in the field, where English is frequently taken to be the source language.

For the English word vector embedding, we used the pre-trained model made publicly available by the researchers for GloVe on their website at http://nlp.stanford.edu/projects/glove/. For French, however, a pre-trained embedding was not conveniently available, so a model was built using GloVe as part of this work, trained using the entire corpus of French-language Wikipedia. Although this did provide a workable word embedding for use in the translation model, the size of the training set for these French word embeddings was limited to only the half a billion tokens that were available, in contrast with the 42 billion tokens that were used to train the provided English word embeddings. In both cases, the representations that were returned belong to a 300-dimensional vector space.

Training was performed using batched stochastic gradient descent, where each batch contained about 128 sentences. The model was set to have 4 layers, in light of the success observed by Sutskever et. al \cite{sutskever2014sequence}; however, the number of nodes in the present model is set at 500 per layer, in contrast with the 1000 used by the earlier work, in order to speed up training and reduce potential overfitting.

In this work, the focus was on training the neural network to work as an effective classifier, which is the key component of the translation framework laid out in Section~\ref{AutoTranslate}. There were two ways to approach this in the context of automatic translation, as outlined in the following sections.

\subsection{Method 1}
In this training method, the focus is entirely upon building a strong classifier under the belief that everything has gone perfectly up to the current position in the translation. Given a sequence of French input word vectors $(x_{0},\dots,x_{F})$ and the corresponding sequence of \emph{correct} English translation word vectors $(y_{0},\dots,y_{E})$, this method generates predicted translation words $\hat{y_k}$ one at a time, but uses the ground-truth translations instead of its own predictions every time it looks for the next word in the sequence:

\begin{enumerate}
	\item input: $((x_{1},\dots,x_F),())$; output: $\hat{y_{1}}$; expect: $y_1$
    \item input: $((x_{1},\dots,x_F),(y_{1}))$; output: $\hat{y_{2}}$; expect: $y_2$ \\
    \dots
    \item input: $((x_{1},\dots,x_F),(y_{1},\dots,y_{E-1}))$; output: $\hat{y_{E}}$; expect: $y_E$
\end{enumerate}

This model is built upon high quality input and should therefore act a powerful classifier. However, since it is only trained upon flawless input data, it might encounter difficulties in practice if any word happens to be mistranslated partway through a sequence, as the model does not have experience correcting itself from a partially incorrect translation sequence.

\subsection{Method 2}
The focus of the second training method is to build a classifier that is able to handle robustly correct itself from intermediate translation errors. As in the previous approach, this method generates predicted translation words $\hat{y_k}$ one at a time, but now uses the model's own predicted translation, rather than the ground-truth, to predict the next translated word in the output sequence:

\begin{enumerate}
	\item input: $((x_{1},\dots,x_F),())$; output: $\hat{y_{1}}$; expect: $y_1$
    \item input: $((x_{1},\dots,x_F),(\hat{y_{1}}))$; output: $\hat{y_{2}}$; expect: $y_2$ \\
    \dots
    \item input: $((x_{1},\dots,x_F),(\hat{y_{1}},\dots,\hat{y_{E-1}}))$; output: $\hat{y_{E}}$; expect: $y_E$
\end{enumerate}

This model is much more practical and robust than the previous one, as it is trained on its own output, much like in the situations where the model is applied in practice; as a result, there should be a much smaller difference between test error and training error, and the model will perform better in real applications.

The major disadvantage with this approach is that training is a much more difficult and time-consuming task. For much of the early part of training, the output predicted words will be close to random and the neural network will have to learn how to use this random noise to predict the next correct word. Not only would each step in the prediction process be slower during the training period, but the required number of training iterations would rise by orders of magnitude, compared with the optimistic approach of Method 1.

\subsection{Technical Details}
Some of the specific decisions made regarding the implementation of model training for this work are described below.

\begin{enumerate}
  \item Intial weights were set using the Glorot initialization \cite{glorot2010understanding} with weights sampled from the uniform distribution. This is designed to set weights that are neither too small nor too large, so as to avoid vanishing or exploding signals.
  \item We did not expect many issues with vanishing gradients, since the GRU-based model is generally able to safely handle this problem. To avoid the unlikely risk of exploding gradients, a hard clip for gradient values was kept at 100.
  \item The entire collection of batches is shuffled at the start of every training epoch. Training then proceeds stochastically through the sequence of batches. There were about 6000 batches in total and a single epoch took 5-6 hours using a single Nvidia 980 Ti processor.
  \item The training error for each word prediction in the sequence was computed by taking the cross entropy between the predicted vector of conditional probabilities vector produced by the neural network, and the ground-truth vector which has a value of 1 at the index of the correct next word and 0 for all other indices.
  \item Nesterov momentum \cite{sutskever2013importance} was used during the gradient descent process in an attempt to speed up the training procedure.
\end{enumerate}

\section{Results}
\label{Results}

In general, the trained neural network for this work did not perform as well as other state-of-the-art models, but was able to achieve a modest success considering the constrained time and resources that were available.\footnote{It is unfortunate that despite repeated and earnest consultation with Sharcnet support services, technical issues prevented the research team from obtaining the full benefit of Sharcnet resources in time to complete this work.} Of the two methods described in Section~\ref{TrainMethods}, the second method is believed to provide a stronger and more robust translation model. However, in practice, the training procedure was prohibitively slow as the model could not identify useful patterns in translation quickly from the original randomly-generated translations. As a result, only the first method was successfully implemented and tested.

The results of the translation model can be qualitatively evaluated by looking at some of the automatically translated sentences, and comparing them with the correct English translations. A sample of some translations can be seen in Table 1. For comparison, the translation offered by Google Translate is also provided.

Through the present model, translation quality quickly deteriorates for longer sentences, even those within the training set. The results on some of the shorter sentences, on the other hand, are reasonable and sometimes even creative. However, it is clear that this model is not powerful enough to fully translate by itself, achieving a validation test score of only 22.9\% -- that is, given an input sequence of words in French and a partial sequence of words in the English translation so far, the probability of identifying the next word in the translation correctly is a little more than 1 in 5.

A possible reason for the struggling performance of this model overall is that the neural network was trained to estimate the conditional probability given a correct sequence of inputs, but the translation process builds upon predicted output words that may not be correct. Once the predicted word sequence goes ``off track'', it no longer has a good context to translate from, and having never learned how to go back to the correct sequence, the model will understandably face difficulty. It should be noted that if the model is asked to predict the next word when given an input sequence of words in French and a partial sequence of words in the \emph{correct} English translation, the probability of identifying the next word in the translation correctly is nearly doubled, at about 40.3\%.\footnote{Perhaps 40\% is not particularly good either, but machine translation is a difficult problem, and it is comforting to note that even Google Translate has issues with translating some sentences as well.}

\section{Recommendations and Conclusions}
\label{Conclusions}

The GRU-based model presented in this work showed early signs of potential for leading to a powerful automatic translator. Although the resulting translations were not always of the highest quality, there is almost always a visible link to the intended translation, and there is frequently some coherency within the overall prediction. As a starting attempt at a new approach with tight constraints on time and resources, the present success may even be considered quite promising.

In future work, it would be interesting to try using a larger training set, more nodes per hidden layer in the neural network, and a larger dataset for training the French language word vectors. These adjustments would certainly amplify the time required for training, but it is possible that they may lead to a more capable translator. Based on the solitary model examined as part of this work, it not clear what the effects that many of the parameters, such as number of nodes and the size of the vocabulary, would have on the ability of the resulting translator. This would certainly present an interesting avenue of future research.

\section{Acknowledgments}
The authors would like to extend grateful acknowledgments to Dr.~Ghodsi for teaching the course and creating an environment of enthusiasm for learning, both for the students and our algorithms. Thank you as well for taking the time to mark our final projects during the winter holiday.

\small
\begin{center}
\label{ResultsTable}
    \begin{tabular}{ p{4cm} | p{4cm} | p{4cm} | p{2cm}}
    \bf English Translated Sentence & \bf English Correct Sentence & \bf Google Translate & \bf Set \\ \hline
    am much more space . 1 & i ve got plenty of room & I have a lot of space. & Test \\ \hline 
    few things are being used for the journey & a light raincoat is ideal for the trip & a light raincoat is ideal for travel. & Test \\ \hline
    am delighted to see our next meeting & i look forward to our next meeting & I look forward to our next meeting. & Test \\ \hline
    president on a point of order & madam president on a point of order & Madam Chair c is a procedural motion. & Training \\ \hline
    you send a doctor to a doctor & will you send someone go get a doctor & 
are you send someone to fetch a doctor. & Test \\ \hline
    please please call for this silence & please rise then for this minute silence & I invite you to stand for this minute of silence. & Training \\ \hline
    he was born it was very difficult to speak . NF & since he was dressed in black he looked like a priest & as he was dressed in black he looked like a priest. & Test \\ \hline
    will have a painful job but he did not have his surprise & she told him a joke but he didnt think it was funny & she told him a joke but he did not find it funny. & Test \\ \hline
    have had a debate on this subject in the next part session in the debate . NF . NF . & you have requested a debate on this subject in the course of the next few days during this part session & you wanted a debate on this in the coming days during this session. & Training \\ \hline
    declare resumed the session of the european parliament which was held on thursday 19 october which i have had to say and i am sure you will confirm that my meeting was good news . & i declare resumed the session of the european parliament adjourned on friday 17 december 1999 and i would like once again to wish you a happy new year in the hope that you enjoyed a pleasant festive period & resumed session
I declare resumed the session of the European Parliament adjourned on Friday 17 December and I renew my vows in the hope that you had a good holiday.& Training \\ \hline   
    \end{tabular}
    \\[2pt]
    \textbf{Table 1:} Translated phrases obtained using our model, in contrast with the correct translation and the result obtained from Google Translate. ``NF'' implies that the model predicted a word outside of the vocabulary, or a potential sentence break.
\end{center}

\bibliography{refrence}{}
\end{document}